\documentclass[letterpaper, 10 pt, conference]{ieeeconf}  % Comment this line out if you need a4paper

\IEEEoverridecommandlockouts                              % This command is only needed if 
\usepackage{graphics} % for pdf, bitmapped graphics files
\usepackage{epsfig} % for postscript graphics files
\usepackage{mathptmx} % assumes new font selection scheme installed
\usepackage{times} % assumes new font selection scheme installed
\usepackage{amsmath} % assumes amsmath package installed
\usepackage{amssymb}  % assumes amsmath package installed
\usepackage{tikz}
\usetikzlibrary{shapes.geometric, arrows, arrows.meta, patterns, calc, positioning, fit, decorations.pathreplacing}
\usepackage{adjustbox}
\usepackage{esvect}
\usepackage{makecell}
\usepackage{hyperref}
\usepackage{graphicx} % Allows the use of \begin{figure} and \includegraphics
\usepackage{float} % Useful for specifying the location of a figure ([H] for ex.)
\usepackage{caption} % Adds additional customization for (figure) captions
\usepackage{subcaption} % Needed to create sub-figures
\usepackage{comment}
\usepackage{svg}
\usepackage{multirow}
\usepackage{cleveref}
\title{\LARGE \bf
Synthetic Dataset Generation for Autonomous Mobile Robots Using 3D Gaussian Splatting for Vision Training
}

\author{Aneesh Deogan$^{1,*}$, Wout Beks$^{1}$, Peter Teurlings$^{1}$, Koen de Vos$^{1}$,\\ Mark van den Brand$^{2}$, and René van de Molengraft$^{1}$% <-this % stops a space
\thanks{This work was supported by the AI-MATTERS project co-funded by the European Union under grant agreement number 101100707}% <-this % stops a space
\thanks{$^{1}$Robotics Section in the Department of Mechanical Engineering,
        Eindhoven University of Technology, The Netherlands
        }%
\thanks{$^{2}$ Software Engineering and Technology in the Department of Mathematics and Computer Science,
        Eindhoven University of Technology, The Netherlands
        }%
\thanks{$^{*}$ Corresponding author a.s.deogan@tue.nl
        }%
}

\begin{document}

\maketitle
\thispagestyle{empty}
\pagestyle{empty}

%%%%%%%%%%%%%%%%%%%%%%%%%%%%%%%%%%%%%%%%%%%%%%%%%%%%%%%%%%%%%%%%%%%%%%%%%%%%%%%%
\begin{abstract}

Annotated datasets are critical for training neural networks for object detection, yet their manual creation is time- and labour-intensive, subjective to human error, and often limited in diversity. This challenge is particularly pronounced in the domain of robotics, where diverse and dynamic scenarios further complicate the creation of representative datasets. To address this, we propose a novel method for automatically generating annotated synthetic data in Unreal Engine. Our approach leverages photorealistic 3D Gaussian splats for rapid synthetic data generation. We demonstrate that synthetic datasets can achieve performance comparable to that of real-world datasets while significantly reducing the time required to generate and annotate data. Additionally, combining real-world and synthetic data significantly increases object detection performance by leveraging the quality of real-world images with the easier scalability of synthetic data. To our knowledge, this is the first application of synthetic data for training object detection algorithms in the highly dynamic and varied environment of robot soccer. Validation experiments reveal that a detector trained on synthetic images performs on par with one trained on manually annotated real-world images when tested on robot soccer match scenarios. Our method offers a scalable and comprehensive alternative to traditional dataset creation, eliminating the labour-intensive error-prone manual annotation process. By generating datasets in a simulator where all elements are intrinsically known, we ensure accurate annotations while significantly reducing manual effort, which makes it particularly valuable for robotics applications requiring diverse and scalable training data\footnote{\url{https://gitlab.tue.nl/20181640/synthetic-dataset-generation}}.

% The performance in real-life matches is assessed, validating the method and checking if motion blur should be incorporated into the datasets. 

\end{abstract}

%%%%%%%%%%%%%%%%%%%%%%%%%%%%%%%%%%%%%%%%%%%%%%%%%%%%%%%%%%%%%%%%%%%%%%%%%%%%%%%%
\section{INTRODUCTION}

% Introduce the growing importance of realistic image datasets in fields like computer vision, machine learning, and AI

\begin{figure*}
  \centering
  \includegraphics[width=\textwidth]{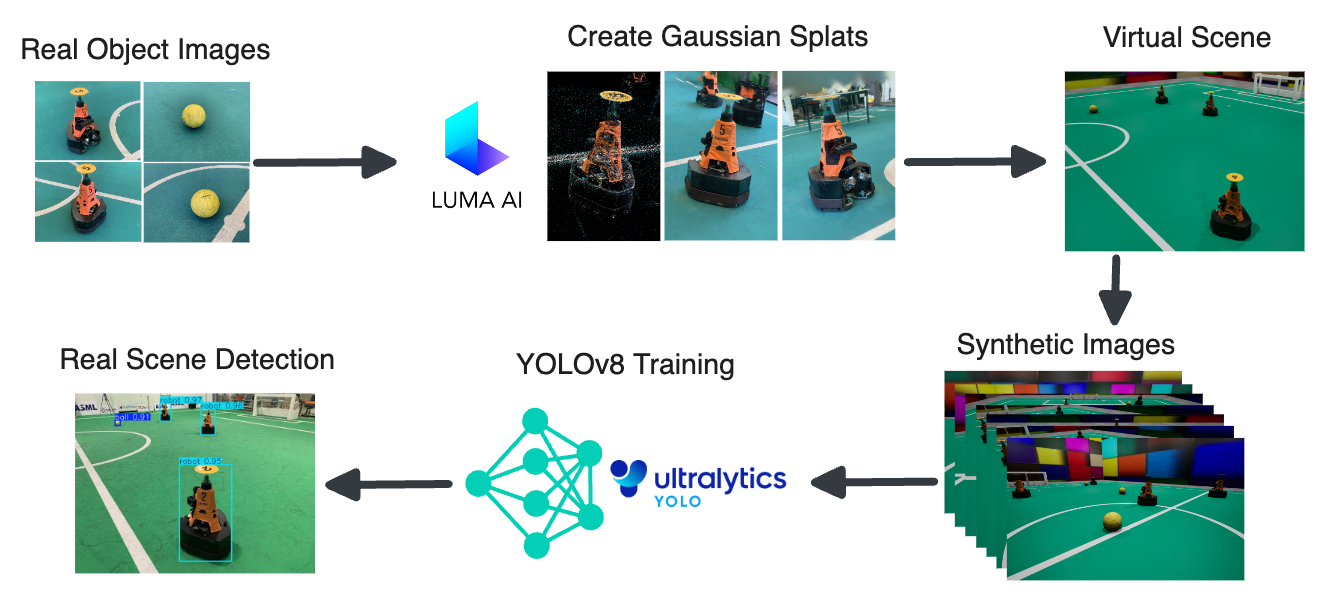}
  \caption{The proposed pipeline for generating synthetic datasets for object detection using 3D Gaussian splats.}
  \label{the_method}
\end{figure*}

% why robotics and computer vision and CNNs
Recent advancements in robotics have highlighted the growing importance of computer vision in enabling robots to perform complex tasks across diverse fields, such as agriculture and industry~\cite{goel2020robotics}. In these environments, robots must manipulate objects in dynamic settings where object positions and orientations cannot be predefined. Computer vision, particularly through deep learning-based methods like convolutional neural networks (CNNs), plays a critical role in enabling robots to perceive and understand their environment~\cite{shreyas20213d}. Supervised CNNs have proven superior to traditional approaches~\cite{o2020deep, janiesch2021machine}, with state-of-the-art models such as YOLO (You Only Look Once) \cite{redmon2016you} demonstrating how advanced vision techniques empower robots to perceive and interact with their environment effectively~\cite{sultana2020review}.

% data gathering/labelling
Deep CNNs require large annotated datasets for training, testing, and operation~\cite{sun2017revisiting}. Annotations are a ground truth for supervised model training and evaluation, which involves manually marking bounding boxes for each object. Annotating real images is labour-intensive, time-consuming, and subjective to human errors in inaccurately marking the bounding boxes. This hinders the practical implementation and adoption of modern CNN-based computer vision systems for robotic applications.

Robust object detection requires large, diverse, and balanced datasets~\cite{salari2022object}, but real-world data collection is costly and often lacks diversity. Collecting data for short periods of time reduces costs but results in smaller datasets that fail to capture critical environmental variations such as lighting changes, occlusion, and motion blur, leading to poorer CNN performance. Extending collection periods to address these issues increases costs due to the need for manual annotations, creating a persistent challenge in balancing time, cost, and dataset quality.

% available datasets are limited
Open-source datasets with millions of annotated images are available, eliminating the need for manual data collection and annotation \cite{lin2014microsoft}. However, generic datasets often lack situational relevance \cite{zendel2017analyzing, block2022image}, as they might not include application-specific objects or relevant environmental variations such as unique occlusions, lighting conditions, and dynamic elements like moving objects. These limitations result in suboptimal CNN performance compared to tailored, application-specific datasets. This underscores the need for methods that generate comprehensive and representative datasets.

% why sythetic datasets are needed in AI vision for robotics
Synthetic dataset generation is emerging as a powerful solution to create large and diverse datasets~\cite{paulin2023review}. Although synthetic images may not perfectly replicate real-world scenes, generating data in virtual environments with explicitly defined object positions and camera parameters enables an efficient and cost-effective approach. Since the generator inherently knows every element of the image at the pixel level, it can automatically produce accurate, instance-level annotations \cite{taylor2007ovvv}. This eliminates human errors common in manual annotation, such as fatigue-induced inaccuracies from misaligned bounding boxes, ensuring consistent and precise annotations. Additionally, synthetic data enables controlled variation of coefficients, variables, and parameters, fostering data diversity and more robust, generalizable models. These benefits have driven the use of synthetic data in robotics applications, including agriculture \cite{saraceni2024self}, autonomous driving \cite{song2023synthetic}, UAVs \cite{UAV_motion_blur}, and manufacturing \cite{block2022image}.

%%%%%%%%%%%%%%%%%%%%%%%%%%%%%%%%%%%%%%%%%%%%%%%%%%%%%%%%%%%
% INCLUDE REFERENCE HERE: smth like - there exist datasets that provide a wide range of photorealistic models. 
% - how are these models created? Nerfs? are 3DGS better?
%%%%%%%%%%%%%%%%%%%%%%%%%%%%%%%%%%%%%%%%%%%%%%%%%%%%%%%%%%%
To minimise the domain gap between real and synthetic images, the synthetic data must be visually realistic \cite{rozantsev2015rendering}. Achieving this requires photorealistic object models, such as those in open-source datasets like Google Scanned Objects (GSO) \cite{downs2022google}. However, these datasets often lack application-specific objects and are slow to generate large volumes of diverse synthetic images due to the high polygon density of the models. This highlights the need for photorealistic object models capable of real-time rendering, enabling rapid generation of large-scale, diverse synthetic datasets.

% nerfs and GANs are quicker but ... therefore, 3dgs's

% why sythetic datasets with composed objects (via pictures) are needed in AI vision for robotics
Tools such as Generative Adversarial Networks (GAN)~\cite{goodfellow2014generative}, Neural Radiance Fields (NeRFs)~\cite{Nerf}, or 3D Gaussian Splats (3DGS)~\cite{3d_gaus_splat} can generate photorealistic models capable of real-time rendering. Although these models may not match the photorealism of GSO models, they are significantly faster in generating diverse and representative large-scale synthetic datasets. Since data diversity is more critical than marginal gains in photorealism~\cite{nowruzi2019much}, these tools offer a practical balance for photorealistic and scalable synthetic data generation.

% why gaussian splats is needed in sythetic datasets with composed objects (via pictures) in AI vision for robotics
3DGS's offer significant advantages over GANs and NeRFs for synthetic data generation. GANs struggle with multi-view inconsistency, ignoring the underlying 3D geometry of an object during generation. Although NeRFs overcome this issue, they are hindered by slow training and rendering speeds, even with recent efforts to accelerate them \cite{yariv2023bakedsdf}. NeRFs still require powerful GPUs and are impractical for real-time applications on consumer devices. In contrast, 3DGS's rasterise Gaussian ellipsoids to approximate 3D scenes, achieving comparable view-synthesis quality while enabling fast convergence and real-time rendering. By abandoning neural networks and directly optimising Gaussian ellipsoids, 3DGS's ensure efficient training and rendering, making it ideal for rapid diverse large-scale synthetic data generation. This has enabled its application in the generation of synthetic data for fields such as medical imaging \cite{zeng2024realistic}.

% more variability in datasets using 3DGS:
% All these advantages have led researchers to use 3DGS's, for example in synthetically generating medical image datasets \cite{zeng2024realistic}. To allow more tractable control over texture and lighting, researchers have started to disentangle texture and lighting to enable independent editing \cite{gao2023relightable}. High-quality reconstruction and photorealistic rendering of dynamic scenes, allowing the user the ability to add variations to the dataset by deforming the 3DGS \cite{yang2024deformable}.

%%%%%%%%%%%%%%%%%%%%%%%%%%%%%%%%%%%%%%%%%%%%%%%%%%%%%%%%%%%%
%%%%%%%%%%%%%%%%%%%%%%%%%%%%%%%%%%%%%%%%%%%%%%%%%%%%%%%%%%%%
% You go from '3DGS are good' to 'application robot soccer'
%
% - how about explain different applications of synthetic data (agriculture, etc.)
% - OR explain how soccer robots are a good application and how it relates to robotics in the wider sense (basically why robocup is good for industry robotics)
% - OR do what Rene said: include a short paragraph of robocup MSL at the start of the methods section (application area, like in the SPG paper)
%%%%%%%%%%%%%%%%%%%%%%%%%%%%%%%%%%%%%%%%%%%%%%%%%%%%%%%%%%%%
%%%%%%%%%%%%%%%%%%%%%%%%%%%%%%%%%%%%%%%%%%%%%%%%%%%%%%%%%%%%

% main contribution (main gap in the literature that this paper adresses) (% Briefly mention the key contributions of your work, such as novel methodologies, comparative analyses, or improvements in dataset realism.)
The main contributions of this paper are leveraging 3DGS's for the rapid generation of representative synthetic datasets for highly dynamic, uncertain, and rapidly changing robotic applications, demonstrated through its applicability in the dynamic and varied context of robot soccer. This method can be used to automate the generation of annotated datasets and to improve performance when an existing real-world-labelled dataset is available, but limited in diversity and size. %Furthermore, this method can be used as an input to Segment Anything Models (SAM2) for the segmentation process. % with other general purpose platforms like Unity \cite{borkman2021unity} or NVIDIA Omniverse Isaac Sim \cite{singh2024synthetica}.

% Summarize the structure of the paper, briefly mentioning what each section covers.

% \section{RELATED WORK}

% Limitations of current dataset generation techniques (e.g., synthetic datasets, labor-intensive annotations, and lack of diversity).

% What is 3D Gaussian Splatting and why it has become a promising technique for creating realistic 3D models.

% Connect how 3D Gaussian Splatting can be applied to generate realistic image datasets.

% Discuss why traditional 3D rendering techniques or other synthetic dataset generation methods fall short in terms of realism, scalability, or computational efficiency.

% Clearly outline the primary goal of your paper—how you use 3D Gaussian Splatting for generating realistic image datasets.
% Robots are rarely stationary, so they must be robust against motion blur.
% While self-supervised synthetically blurred scenes is an option \cite{alvarez2019self}, Unreal Engine uses velocity geometry buffer which provides accurate motion blur in complex scenarios where objects and the camera are moving simultaneously \cite{shannon2017unreal}.

% Takes motion blur and occlusion into account in their synthetic datasets \cite{huh2018simple}.

% But we don't need to model motion blur because the effects are minimal. 

\section{APPLICATION AREA}

Robotic soccer is chosen as a benchmark due to its highly dynamic and uncertain environment \cite{robocup}. In RoboCup MSL, teams of five fully autonomous robots compete under FIFA-like rules, with all sensors and computing modules required to be onboard. While intra-team communication via Wi-Fi is allowed, all other objects (e.g., ball, opponents) must be detected using onboard sensors.

The constant movement of the ball, teammates, and opponents creates a rapidly changing environment, making future state predictions challenging. Thus, real-time multi-object detection is critical for success in such scenarios.

\section{METHODS}

Figure \ref{the_method} illustrates the proposed method. By capturing images of the target objects, we create 3D photorealistic models. These models are then integrated into a virtual environment to generate representative synthetic datasets. Using these annotated datasets as input for supervised CNN object detection models enables the rapid development of a robust detection solution.

\subsection{Object Modelling}

\begin{figure}[h]
    \centering
  \begin{subfigure}[t]{0.49\linewidth}
    \includegraphics[width=\linewidth]{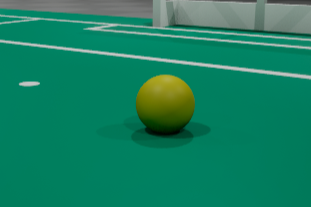}
    \vspace{-4.2mm}
    \caption{Simulated Ball.}
    \label{sim_ball}
  \end{subfigure}
  \begin{subfigure}[t]{0.49\linewidth}
    \includegraphics[width=\linewidth]{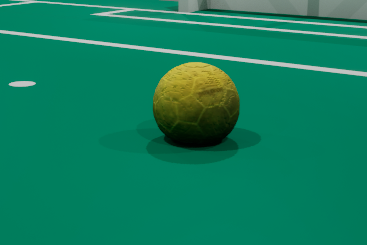}
    \caption{3D Gaussian Splat Ball.}
    \label{3dgs_ball}
  \end{subfigure}
    \caption{The (a) simulated and (b) 3D Gaussian splat balls used in the virtual environment for synthetic data generation.}
    \label{simple_object_models}
\end{figure}

\begin{figure}[h]
  \begin{subfigure}[b]{0.32\linewidth}
    \includegraphics[width=\linewidth]{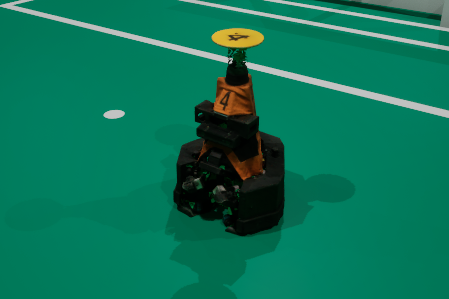}
    \caption{TechUnited~\cite{techUnited}.}
    \label{3dgs_robot_techunited}
  \end{subfigure}
  \hfill
  \begin{subfigure}[b]{0.32\linewidth}
    \includegraphics[width=\linewidth]{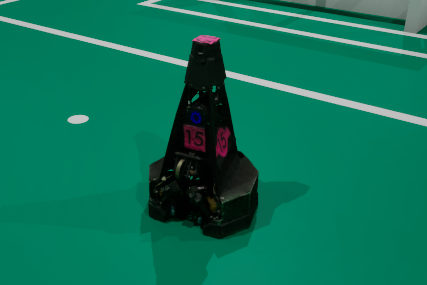}
    \caption{Falcons 1~\cite{falcons}.}
    \label{3dgs_robot_falcon1}
  \end{subfigure}
  \hfill
  \begin{subfigure}[b]{0.32\linewidth}
    \includegraphics[width=\linewidth]{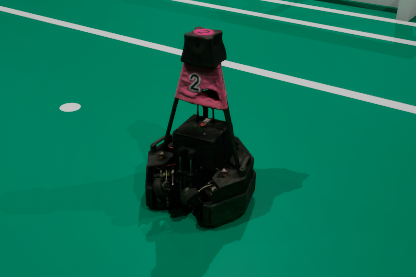}
    \caption{Falcons 2~\cite{falcons}.}
    \label{3dgs_robot_falcon2}
  \end{subfigure}
    \caption{3D Gaussian splats of different robots used in the virtual environment for synthetic data generation.}
    \label{complex_object_models}
\end{figure}

Figures \ref{simple_object_models}-\ref{complex_object_models} illustrate the modelled objects. Figure \ref{sim_ball} is a low-fidelity simulated model of the ball created in Unreal Engine, while Figures \ref{3dgs_ball}-\ref{complex_object_models} show the 3DGS models of the ball and different robots. Coarse simulated objects offer the advantage of being easy to create in any simulation environment—for instance, a yellow sphere can represent a ball. In contrast, 3DGS models deliver photorealistic representations more efficiently than highly realistic simulated objects. 

3D Gaussian splatting is a rasterisation technique that allows real-time rendering of photorealistic scenes learned from small samples of images.  First the Structure from Motion (SfM) method \cite{schonberger2016structure} is used to estimate a point cloud from a set of images. Next, each point is converted to a Gaussian. The training procedure uses Stochastic Gradient Descent. This training includes the following steps. Each Gaussian is projected in 2D from the camera perspective. The Gaussians are then sorted by depth. For each pixel, it iterates over each Gaussian front-to-back, blending them together. Then, the loss based on the difference between the rasterised image and ground truth image is calculated, on which the Gaussian parameters are adjusted. Finally, automated densification and pruning are applied \cite{3d_gaus_splat}.

Processing each 3DGS took approximately an hour on average, resulting in highly realistic models, shown in Figures \ref{3dgs_robot_techunited} through \ref{3dgs_robot_falcon2}. These models were produced by capturing images with a standard smartphone camera (iPhone XR) and processed by LUMA AI \cite{lumaAI_gaussian_splats} which generated the 3DGS's. %For enhanced security, 3DGS models can also be generated on a local network.
%`
\subsection{Synthetic Dataset Generation}
\begin{figure}[h]
    \centering
    \includegraphics[width=0.85\linewidth]{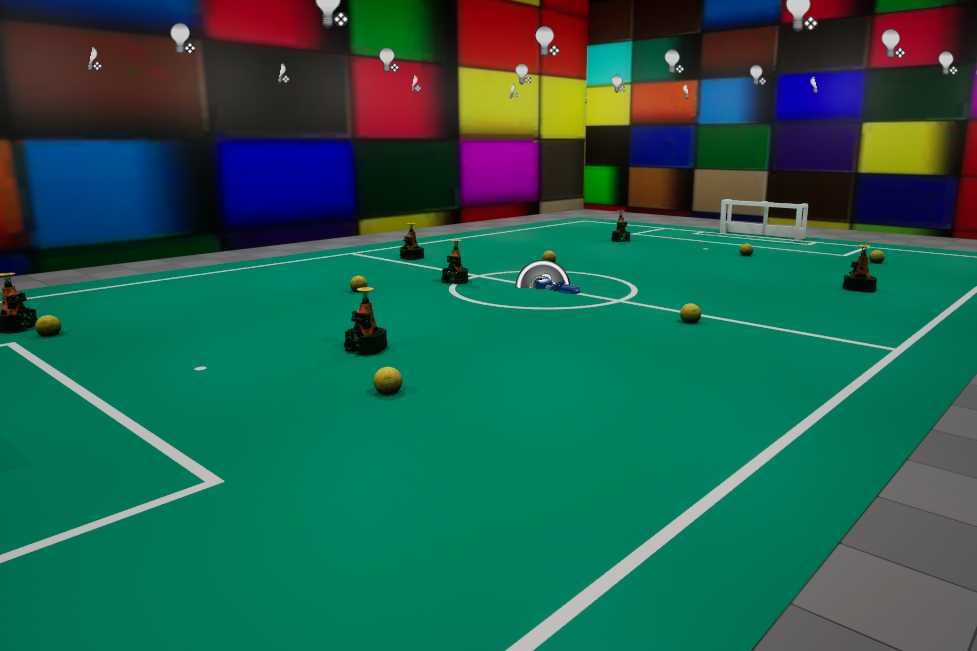}
    \caption{Example of the virtual environment, including the field and goals, the lights, the background augmentation, and 3D Gaussian splats of balls and robots.}
    \label{virtual_environment}
\end{figure}

Figure \ref{virtual_environment} illustrates the virtual environment used to generate synthetic images. To create representative synthetic datasets, certain factors must closely align with real-world conditions. For example, the camera in Unreal Engine replicates the lens parameters and Field of View (FOV) of the ZED2 camera from StereoLabs, used on the physical robots, with a 110° horizontal, 70° vertical, and 120° diagonal FOV. In addition, the lighting setup mimics the indoor conditions of RoboCup arenas, which feature multiple ceiling-mounted light sources. Background enhancement, as seen in Figure \ref{virtual_environment}, prevents the CNN model from overfitting to a default or empty simulation background.

The simulated environment also facilitates the introduction of domain randomisations or variations in the images of the dataset. Relevant variables include light intensity, camera orientation, ball orientation, and robot orientation. The extent of these variations and their distributions are application-specific. For example, the positions of balls, robots, and cameras are constrained to the field, where we apply a uniform distribution to these variations. This ensures that the distance between objects, such as the robot and the camera, is evenly varied within predefined bounds.

Bounding box calculations can be automated in a virtual environment because the exact orientations of objects and the camera are explicitly known. This allows 3D object coordinates to be projected onto the 2D image plane, where the minimum and maximum extents in screen space define the bounding boxes \cite{bounding_box_plugin}. In real-world images, such automation is not feasible, since object and camera orientations are not explicitly known, requiring manual annotations instead. Using 3DGS models in the virtual environment, generating one annotated synthetic image takes approximately one second on a standard laptop (Intel i7-10750H @ 2.6 GHz, 16GB RAM). This drastically accelerates the annotation process compared to the labour-intensive and time-consuming task of manually annotating real-world images for (re)training CNNs.
\subsection{Validation} 
\label{sec:validation}

YOLOv8, developed by Ultralytics \cite{jocher_ultralytics_YOLO_2023}, is the state-of-the-art object detection algorithm used for the experiments. All trainings are executed on a workstation equipped with a NVIDIA GeForce RTX 4080 GPU, Intel i7-14700K, and 32GB RAM.

% how we train the YOLO
The ball model trainings involved 10 epochs, a batch size of 16, and approximately 1000 images without Automatic Mixed Precision (AMP) training. Default parameters provided by Ultralytics \cite{jocher_ultralytics_YOLO_2023} were used: an initial and final learning rate of $1 \times 10^{-2}$, a momentum of 0.937, and a weight decay of $5 \times 10^{-4}$. Training times are heavily dependent on the workstation, training parameters, the version of YOLO, and the number of images used.

% compare ball models
To assess the feasibility of 3DGS models, a synthetic 3DGS ball dataset (Fig. \ref{3dgs_ball}) was compared to a real-world human-annotated ball dataset. The validation used 70 real hand-annotated images of static and moving balls on a soccer field. A coarse simulated ball model (Fig. \ref{sim_ball}) was also included for comparison, to verify whether non-photorealistic models can suffice for basic objects, such as a yellow sphere representing the ball. Furthermore, pre-trained CNN models can accelerate the annotation process of known objects (a ball), but are ineffective for unknown objects not available in existing datasets (soccer robots). 

% compare robot models
We evaluated the trade-off between annotation speed and detection performance using two synthetic 3DGS datasets—3DGS-2300 (2300 images) and 3DGS-6000 (6000 images)—alongside a 2300-image human-annotated real-world dataset. To leverage the strengths of both, we also created a combined dataset (2000 real-world and 4000 synthetic images). All datasets were trained over 150 epochs. Validation on 70 hand-annotated real-world images provided a comprehensive comparison between each approach.% demonstrated the efficiency and adaptability of the 3DGS approach.

%%%%%%%%%%%%%%%%%%%%%%%%%%%%%%%%%%%%%%%%%%%%%%%%%%%%%%%%%%%%%%%%%%%%%%%%
% EXPLAIN SYNTHETIC IS SO MUCH QUICKER PRODUCING COMPARABLE RESULTS TO REAL-WORLD ANNOTATED DATASETS - AND SINCE THE AMOUNT OF DATA REALLY AFFECTS THE QUALITY OF THE YOLO NETWORK, THE QUICKER YOU CAN PRODUCE MANY DOMAIN RANDOMISED IMAGES, THE QUICKER YOU CAN GET AN ACCEPTABLE YOLO OBJECT DETECTION ALGORITHM. SO IF YOU WANT TO ADD NEW OBJECTS INTO YOU DETECTION ALGORITHM, IT'S MUCH QUICKER TO GENERATE AND THEREFORE HAVE A TRAINED YOLO.
%%%%%%%%%%%%%%%%%%%%%%%%%%%%%%%%%%%%%%%%%%%%%%%%%%%%%%%%%%%%%%%%%%%%%%%%

% complete 3DGS model
Finally, the effectiveness of the 3DGS approach was tested in a match-play context. A synthetic dataset comprising 5000 images and trained over 150 epochs (referred to as 3DGS-All in Table \ref{3dgs_all}) was validated using 85 hand-annotated images from a real-world robot soccer match. This dataset included 4 classes: 3DGS models of the ball and multiple robots (from Fig. \ref{3dgs_ball} and Fig. \ref{complex_object_models}). This evaluation demonstrates the potential of 3DGS-based datasets in RoboCup scenarios. 

% validation metrics
The approach is validated using the conventional metrics: Precision, Recall, F1-Score, Intersection over Union (IoU), and mean Average Precision (mAP) \cite{padilla2020survey}. Precision determines, based on all the detections, how many detections are correct. Recall (or sensitivity) is a measure of the model’s ability to identify all instances of objects present in the images. The F1-Score is derived from the harmonic mean of precision and recall, providing a balanced measure of a model's performance:
\begin{equation}
    \textit{F1-Score}=2\cdot\frac{Precision\ \cdot\ Recall}{Precision\ +\ Recall}.
    \label{f1_score_eq}    
\end{equation}
The mAP is the mean of the Average Precision (AP), which is the area under the precision-recall curve, a plot of precision against recall at different confidence thresholds. It reflects how well a model is performing across different object classes:
\begin{equation}
    mAP = \frac{1}{N} \sum_{i=1}^{N} AP_i,
\end{equation}
where $N$ is the number of classes and $AP_i$ is the average precision for the $i^{th}$ class. In this case, the classes are the ball, Tech United robots~\cite{techUnited}, and Falcons robots 1 and 2~\cite{falcons}. The mAP50 is calculated using an IoU threshold of 0.5, reflecting the model's performance on relatively simple detections, since the IoU) threshold determines how much overlap is required between the predicted bounding box and the ground truth for a detection to be considered a true positive. In comparison, mAP50-95 is determined across ten IoU thresholds ranging from 0.5 to 0.95, with increments of 0.05.

\section{Results}

\subsection{Ball Detection Validation}

\begin{table}[htp]
    \centering
    \addtolength{\tabcolsep}{-0.1em}
    \scriptsize
    \caption{Comparison of a YOLO ball detection for a 3D simulated ball dataset, a 3DGS ball dataset, and a real-world dataset, all validated on hand-annotated real-world images.}
    \begin{tabular}{|c|c|c|c|c|c|}
        \hline
        \textbf{Ball Dataset} & \textbf{mAP50$\uparrow$} & \textbf{mAP50-95$\uparrow$} & \textbf{Precision$\uparrow$} & \textbf{Recall$\uparrow$} & \textbf{F1-score$\uparrow$} \\
        \hline
        \textbf{Simulated} & 0.962 & 0.8888 & 0.938 & 0.944 & 0.941 \\ \hline
        \textbf{3DGS-Ball} & 0.990 & 0.9265 & 0.957 & 0.968 & 0.962 \\
        \hline
        \textbf{Real-World} & 0.994 & 0.9647 & 0.978 & 0.992 & 0.985 \\ 
        \hline
    \end{tabular}
\label{result_ball_validation}
\end{table}

Table \ref{result_ball_validation} presents a comparison of the annotated ball datasets used to train the YOLOv8 object detection algorithm: the low-fidelity simulated ball model dataset (Simulated), the 3DGS dataset (3DGS-Ball), and the baseline real-world dataset comprised of human-annotated images (Real-World). Performance was quantified using the conventional metrics described in Section~\ref{sec:validation}.

The simulated ball dataset demonstrated strong precision and recall but underperformed compared to the real-world dataset. However, the simulated method remains suitable for many applications, achieving an mAP50 of $0.962$ and an F1-score of $0.941$. These results are similar to those reported in \cite{ball_comparison_2}, where a YOLOv3 detection algorithm trained on human-annotated real-world images for soccer robots achieved a recall of $0.92$, precision of $0.92$, and an mAP of 0.871.

The 3DGS dataset slightly outperforms the simulated dataset but still falls short of the real-world dataset. For applications requiring higher accuracy, the 3DGS or real-world datasets are recommended over the simulated dataset. In cases where time constraints make annotating real-world images impractical, the 3DGS method is particularly advantageous. It enables the efficient generation of large and representative annotated datasets, offering greater accuracy than the low-fidelity model, while being faster than relying on human-annotated real-world datasets.

Although the 3DGS method does not fully match the performance of a human-annotated real-world dataset for the same dataset size, it is a viable approach for precision-critical use cases. The mAP50 difference of $0.4\%$ underscores a compelling trade-off of similar performance, with significant time savings by generating one annotated image in 1 second versus 5 seconds plus data collection in real-world scenarios. The $2.39\%$ relative difference in F1-score is largely due to the lower recall, which can be easily increased by generating more images.

In summary, Table \ref{result_ball_validation} suggests that for simple objects, simulated models suffice for the generation of synthetic images. For high-stakes or precision-critical applications, the 3DGS method is a time-efficient choice. However, the human-annotated real-world dataset remains the gold standard.

\subsection{Robot Detection Validation}

\begin{table}[htp]
    \centering
    \addtolength{\tabcolsep}{-0.15em}
    \scriptsize
    \caption{Validating the complex object detection (Tech United robot~\cite{techUnited}) on hand-annotated real-world images.}
    \begin{tabular}{|c|c|c|c|c|c|}
        \hline
        \textbf{Robot Dataset} & \textbf{mAP50$\uparrow$} & \textbf{mAP50-95$\uparrow$} & \textbf{Precision$\uparrow$} & \textbf{Recall$\uparrow$} & \textbf{F1-score$\uparrow$} \\ \hline
        \textbf{3DGS-2300} & 0.813 & 0.6904 & 0.780 & 0.789 & 0.784 \\ \hline
        \textbf{3DGS-6000} & 0.942 & 0.8124 & 0.927 & 0.836 & 0.879 \\ \hline
        \textbf{Real-World-2300} & 0.973 & 0.7756 & 0.955 & 0.918 & 0.936 \\ \hline
        \textbf{Combined-6000} & 0.992 & 0.8995 & 0.969 & 0.986 & 0.977 \\ \hline
    \end{tabular}
    \label{robot_validation}
\end{table}

Table \ref{robot_validation} compares the performance of both datasets on hand-annotated images of static and moving robots on a soccer field. While the real-world dataset outperforms the 3DGS dataset across all metrics with the same number of images, increasing the size of the 3DGS dataset significantly closes the performance gap, achieving comparable results with significantly less manual effort.

Synthetic 3DGS image generation proved significantly faster than manual annotation, with the 2300 and 6000 3DGS image dataset showcasing its scalability, taking 40 and 100 minutes, respectively. Whereas the real-world dataset required additional time for image collection and 8 hours of manual annotation using labelIMG \cite{labelIMG}. 

Ultimately, a hybrid approach that combines real-world and synthetic data delivers the best results. Real-world data offers high-quality images, while 3DGS enables rapid generation of large-scale datasets including domain randomisation. This combination not only improves the quality and diversity of the dataset, but also improves YOLO object detection, since large, diverse, and balanced datasets generally lead to better model performance \cite{sun2017revisiting}.

\subsection{Match Gameplay Validation}

\begin{table}[htp]
    \centering
    % \addtolength{\tabcolsep}{-0.2em}
    \scriptsize
    \caption{Validating the object detection for the 3DGS generated datasets on real life match gameplay.}
    \begin{tabular}{|c|c|c|c|c|c|}
        \hline
        \textbf{Dataset} & \textbf{mAP50$\uparrow$} & \textbf{mAP50-95$\uparrow$} & \textbf{Precision$\uparrow$} & \textbf{Recall$\uparrow$} & \textbf{F1-score$\uparrow$} \\ \hline
        \textbf{3DGS-All} & 0.929 & 0.7316 & 0.941 & 0.844 & 0.890 \\ \hline
    \end{tabular}
    \label{3dgs_all}
\end{table}

Table \ref{3dgs_all} presents the performance of a synthetic dataset comprised of 5000 images using four classes of 3DGS models: the ball, the Tech United robot~\cite{techUnited}, the Falcons robot 1~\cite{falcons}, and the Falcons robot 2~\cite{falcons}. Validation was conducted on hand-annotated real-world images captured by the ZED2 camera mounted on a robot during a RoboCup-like match, highlighting the method's effectiveness in scenarios with multiple dynamic objects on the field.

The results demonstrate strong performance across all metrics, particularly when compared to the prior performance of 3DGS's for robot detection, which was limited to detecting only a single class (Table \ref{robot_validation}). Although recall scores are relatively lower compared to other metrics, precision is more critical in robot soccer, as false positives—like misidentifying background objects as the ball—can lead to incorrect and costly decisions. Furthermore, existing models can compensate for occasional undetected objects, mitigating this limitation. For systems incorporating such models, including estimation or tracking algorithms already used in soccer robots, the reliance on recall is reduced. Consequently, the 3DGS approach offers an efficient alternative with minimal trade-offs in performance. Direct comparison with human-annotated real-world datasets was infeasible due to the lack of a similarly sized dataset encompassing all four classes. This highlights a critical advantage of 3DGS: its ability to rapidly generate large and diverse datasets, facilitating reliable CNN-based object detection.

% This means that incorporating a new object into an object detection deep CNN becomes a straightforward process. This process is not only faster than collecting and manually labelling real-world data but also achieves comparable performance, since the performance of deep CNNs are based on the quality and volume of the dataset~\cite{sun2017revisiting}. 

%These results are comparable to those reported in \cite{song2023synthetic}, where a YOLOv5 vehicle detection algorithm was trained on synthetic data and evaluated using a camera mounted on an autonomous vehicle. Similarly, they align with the findings of~\cite{block2022image}, which employed a YOLOv5 object detection algorithm in a factory setting. Although the latter study focused solely on static objects, it involved a greater number of classes that encompassed a wider variety of object types.

\begin{figure}[h]
    \centering
    \includegraphics[width=\linewidth]{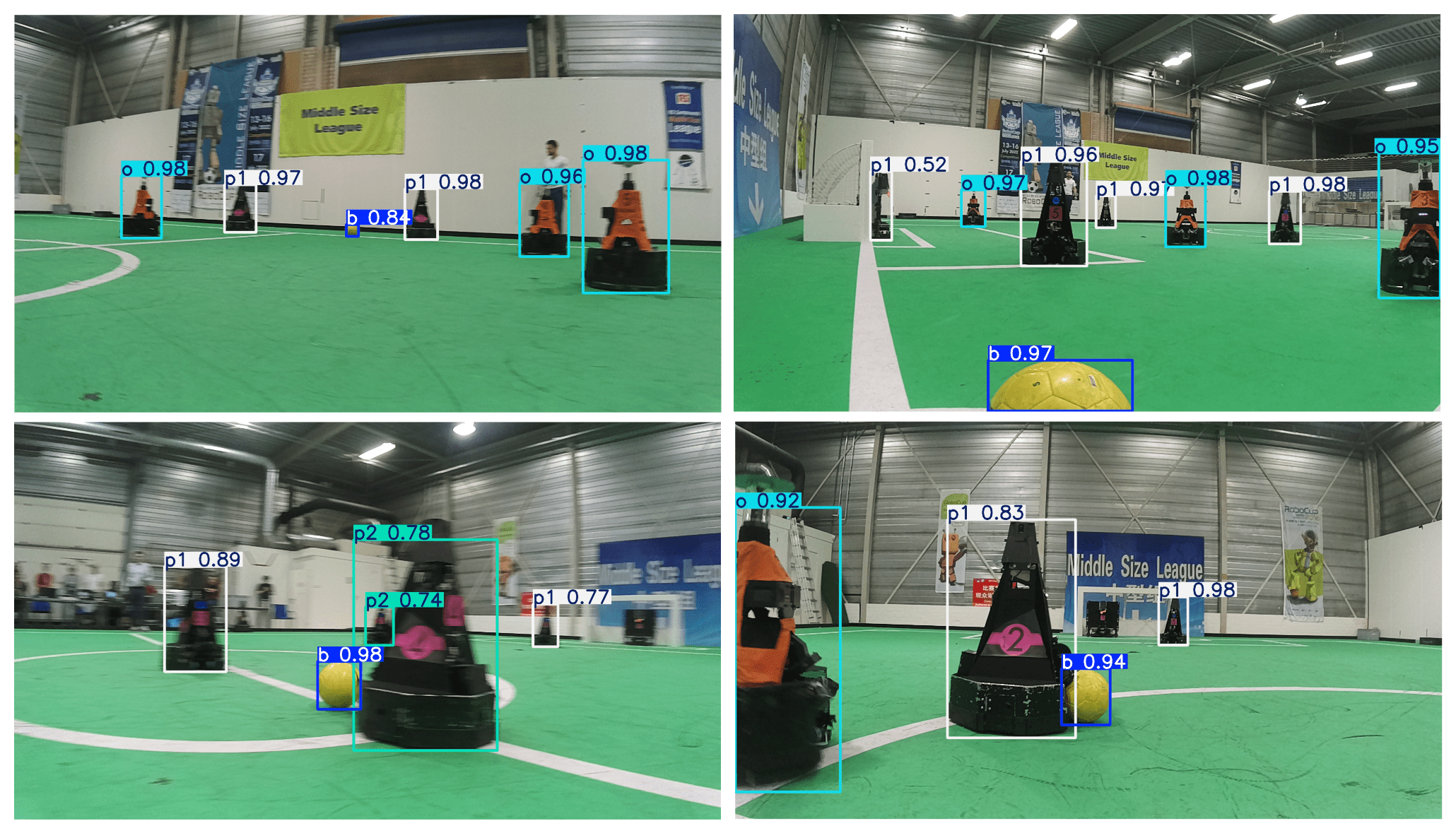}
    \caption{Validation on real life match gameplay. (Ball: b, Tech United robot: o, Falcons robot 1: p1, Falcons robot 2: p2).}
    \label{validation_match}
\end{figure}

The YOLO model trained on synthetic data successfully identifies and detects objects in the scene, visualized through bounding boxes in Figure \ref{validation_match}. It should be noted that goalkeeper robots, which have a different shape compared to outfield soccer robots, were not included in the training datasets. As a result, the goalkeeper robot is not detected in the two lower images of Figure \ref{validation_match}. However, in the top-right image, the goalkeeper is detected, albeit with the low confidence of $0.52$.

This outcome is promising for future iterations, as the detection algorithm demonstrates resilience to slight variations in object appearance, such as damaged or modified covers. This eliminates the need to retrain the model for every minor change, indicating that it does not over-fit to specific robot models. This flexibility underscores the method's potential to adapt to slight variations with minimal impact on performance.

Furthermore, some motion blur is visible in Figure \ref{validation_match}, but it does not appear to influence the detection results. Although extreme motion blur occasionally occurs during matches, these instances are typically brief. Most of the time, the level of motion blur observed does not affect the detection performance. This shows that our 3DGS dataset, which does not include any motion blur, is representative enough to handle RoboCup-like gameplay scenarios effectively.

\section{CONCLUSION}
In conclusion, generating synthetic annotated datasets using 3DGS's has proven to be an effective method for training supervised deep CNNs for object detection in highly dynamic autonomous robotic applications. This approach was effectively applied to robot soccer matches, demonstrating a high mAP50 of $0.929$, offering key advantages such as high precision and rapid dataset generation. Our evaluation showed that, for simple geometric objects, photorealistic models are not necessary with low-fidelity models producing an mAP50 of $0.962$. However, for detecting geometrically complex objects, combining the quality of real-world images with the scalability of synthetic data provides superior performance showing an mAP50 of $0.992$. This hybrid approach offers the precision of real-world datasets and the scalability of synthetic ones, providing representative data for CNNs, even in challenging scenarios like robot soccer.%including motion% blur.

\bibliographystyle{IEEEtran}
\bibliography{references}

\end{document}